\documentclass[11pt]{article}
\usepackage{amsmath,amsfonts}
\usepackage{algorithm}
\usepackage{array}
\usepackage[caption=false,font=normalsize,labelfont=sf,textfont=sf]{subfig}
\usepackage{textcomp}
\usepackage{stfloats}
\usepackage{url}
\usepackage{verbatim}
\usepackage{graphicx}
\usepackage{bm}
\usepackage{hyperref}
\usepackage[capitalize]{cleveref}
\usepackage{svg}
\usepackage{lineno}
\usepackage{enumitem}
\usepackage{breakcites}

\RequirePackage{scrlfile}
\ReplacePackage{scrpage2}{scrlayer-scrpage}

\usepackage[backend=biber, style=authoryear, natbib=true, hyperref=true]{biblatex}
\DeclareDelimFormat[bib,biblist]{nametitledelim}{\addcomma\space}
\DeclareCiteCommand{\cite}[\mkbibbrackets]
  {\usebibmacro{prenote}}
  {\usebibmacro{citeindex}%
   \usebibmacro{cite}}
  {\multicitedelim}
  {\usebibmacro{postnote}}

\usepackage{setspace}
\usepackage{geometry}

\usepackage[singlelinecheck=false]{caption} 

\usepackage[T1]{fontenc}    
\usepackage[british]{babel}       
\usepackage{url}            
\usepackage{booktabs}       
\usepackage{amsfonts}       
\usepackage{nicefrac}       
\usepackage[nopatch=eqnum]{microtype}      
\usepackage{graphicx}
\usepackage{amsmath}
\usepackage{xcolor}
\usepackage{bm}
\usepackage{dsfont}
\usepackage{wrapfig}
\usepackage{float}
\usepackage{capt-of}
\usepackage{caption}

\usepackage{xr}

\usepackage{comment}
\usepackage{algorithm}
\usepackage{algpseudocode}

\usepackage{csquotes}
\algnewcommand\algorithmicforeach{\textbf{for each}}
\algdef{S}[FOR]{ForEach}[1]{\algorithmicforeach\ #1\ \algorithmicdo}

\usepackage[backend=biber, style=authoryear, natbib=true, hyperref=true]{biblatex}
\DeclareDelimFormat[bib,biblist]{nametitledelim}{\addcomma\space}
\DeclareCiteCommand{\cite}[\mkbibbrackets]
  {\usebibmacro{prenote}}
  {\usebibmacro{citeindex}%
   \usebibmacro{cite}}
  {\multicitedelim}
  {\usebibmacro{postnote}}
\usepackage[utf8]{inputenc} 
\begin{document}


\title{How can neuromorphic hardware attain brain-like functional capabilities?}
\author{Wolfgang Maass}
\maketitle

Research on neuromorphic computing is driven by the vision that we can emulate brain-like computing capability, learning capability, and energy-efficiency in novel hardware. Unfortunately, this vision has so far been pursued in a half-hearted manner. Most current neuromorphic hardware (NMHW) employs brain-like spiking neurons instead of standard artificial neurons. This is a good first step, which does improve the energy-efficiency of some computations, see \citep{rao2022long} for one of many examples. But current architectures and training methods for networks of spiking neurons in NMHW are largely copied from artificial neural networks. Hence it is not surprising that they inherit many deficiencies of artificial neural networks, rather than attaining brain-like functional capabilities.

 Of course, the brain is very complex, and we cannot implement all its details in NMHW. Instead, we need to focus on principles that are both easy to implement in NMHW and are likely to support brain-like functionality. The goal of this article is to highlight some of them.
 
Although also other areas of the brain, such as the hippocampus, provide a rich source of inspiration for NMHW with new functional capabilities, I can discuss here only one brain area, and I have chosen the neocortex, a brain area that is central for its computational prowess. The neocortex can be seen in first approximation as a thin sheet of neurons that is structured like a tapestry, i.e., it is stitched together from repeating local circuit modules to which one commonly refers as \textbf{cortical microcircuits (CMs)}, see \citep{douglas2004neuronal, harris_sheperd}. Porting the functionally most relevant design principles of CMs into NMHW is an attractive and feasible target for the next generation of NMHW design.

\vspace{0.5cm}

\textbf{\em Design Principle 1: CMs consist not just of one or two, but over 100 genetically different types of spiking neurons with different computational roles.}

This principle implies that the architecture of CMs is fundamentally different from the randomly connected networks, typically consisting of one or two types of spiking neurons, that are commonly implemented in NMHW. In fact, the sophisticated genetically encoded structure of CMs is more reminiscent of highly structured digital circuits such as CPUs. However, they employ small populations of units, rather than single units, for specific computational roles,
and are therefore more robust to failures of individual units. In fact, CMs have in some sense an even more sophisticated structure than CPUs because they employ a substantially larger repertoire of different units. One prominent example are genetically distinct types of excitatory neurons (pyramidal cells) that report specific prediction errors for top-down predictions of visual flow \citep{o2023molecularly}. 

Another instructive example are different types of inhibitory neurons, which play in CMs a role for computation and learning that is very different from that in current NMHW. There one commonly views inhibitory neurons as clones of excitatory neurons whose outputs have a negative sign. This is convenient for emulating arithmetical computations with negative and positive values, or artificial neural networks. But in the brain, inhibitory neurons play quite different roles. Think for example of populations of specific types of excitatory neurons as being "experts" for specific knowledge or tasks, and inhibitory neurons as controllers which determine which "expert" is allowed to impact the computational task at hand, and which expert is allowed to improve its competence by learning from the current task through synaptic plasticity. According to \citep{harris_sheperd, billeh2020systematic, campagnola2022local} one type of inhibitory neurons, (PV cells), can veto the firing of selected excitatory neurons, which they target through dozens of synapses on their soma. Another inhibitory neuron type (SOM cells) blocks activity and synaptic plasticity in selected input regions (dendrites) of specific excitatory neurons. A 3rd inhibitory neuron type (VIP cells) inhibits both of the previously mentioned types of inhibitory neurons. In other words, VIP cells are disinhibitory: They can remove the inhibitory lock for firing and/or synaptic plasticity for specific populations of excitatory neurons. Their function points to a fundamental difference between CMs on one hand, and current artificial neural networks and NMHW on the other hand: Computation and learning are commonly treated as too distinct processes in NMHW, but are intertwined in sophisticated ways in CMs. Most in-vivo results on
synaptic plasticity point to the involvement of one or several gating factors \citep{chereau2022circuit}that result from the firing of specific populations of neurons. Some of these gating factors, such as disinhibition through VIP cells, is automatically local. But also neuromodulatory gating signals were recently found to be much more target specific than previously assumed. Such a target- and context-specific local regulation of synaptic plasticity is likely to alleviate problems with continual learning that exist in current NMHW, but not in brains.

The genetic code specifies connections probabilities between each pair of the over 100 neuron types \citep{billeh2020systematic, campagnola2022local, chen2022data, chen2023data}. Our analysis \citep{stockl2021structure} shows that this enables the genetic code to program specific computational capabilities into CMs that do not require any learning, as postulated by \citep{zador2019critique}. In fact, in \citep{stockl2021structure} we had not even exploited that different neuron types can have very firing properties, which is likely to enhance innate computing capabilities of CM. Another next step will be to study how desirable learning biases can be programmed with in this way into NMHW, thereby enabling them to learn from few examples.

\vspace{0.5cm}

\textbf{\em Design Principle 2: CMs employ soft rank order coding, i.e., a form of temporal coding, using very sparse activity}.  

The idea to use spike timing for encoding analog values had already been proposed a long time ago. However, rank order coding with single spikes is not robust to deletion or addition of single spikes. The neocortex employs a less brittle rank order coding scheme: Instead of single spikes, the rank order of the times of peak firing activity of different neurons is used to encode information. This type of soft rank order coding, which can be reproduced in CM-like NN models \citep{chen2023data}, is robust to timing jitter, and to the deletion or addition of single spikes. Rank order coding is of particular interest for NMHW because it has recently been shown to provide close-to-optimal energy efficiency for neural coding, see the Supplement.

\vspace{0.5cm}

\textbf{\em Design Principle 3: CMs employ a sophisticated combination of segregation and integration of information over neurons}.

Segregation of information in specialized populations of neurons and the integration of their contribution to a coherent network output has been argued to be characteristic for the organization of brain computations \citep{tononi1998complexity}. The analysis of \citep{chen2023data} shows that this principle can be reproduced in CM models.  Specifically, projection neurons that report network outputs of CMs turn out to be highly sensitive to the firing activity of small sets of "expert neurons". Furthermore, this holds in spite of the remarkable level of noise to which CM-computations are subjected. On the other hand, we found that those spiking and non-spiking neural network architectures that have so far been implemented in NMHW do not have this property \cite{chen2023data}. Note that functional segregation tends to enhance continual learning, since not every neurons gets involved in every computational process.

\vspace{0.5cm}
\textbf{\em Design Principle 4: Computing capabilities of CMs are shaped by diverse local synaptic plasticity rules.}

One commonly implements learning in NMHW with a single rule for synaptic plasticity, such as STDP, or with a single global learning scheme such as BPTT. But the power of STDP for installing computing capabilities in NMHW is quite limited, and BPTT is not suitable for on-chip learning. Hence new learning methods need to be explored for NMHW. The brain suggests to use different plasticity rules for different types of neurons and different forms of learning. Furthermore, most of the biologically found rules do not require a teacher. Many of them do not even depend on postsynaptic firing, and are therefore fundamentally different from STDP and other Hebb-like plasticity rules, see \cite{chereau2022circuit} and further references in the Supplement. The rich repertoire of synaptic plasticity rules in CMs is likely to provide a functionally powerful alternative to BPTT that is suitable for on-chip learning in NMHW. This diverse set of local plasticity rules is also likely to enhance the emergence of neurons that become selective for complex combinations of features in sensory input streams, thereby providing a substantially more sparsely active but still functionally powerful alternative to CNNs for visual processing.

\vspace{0.5cm}

Implementing selected facets of these four design principles in NMHW provides a road map for reproducing in NMHW more energy efficiency through sparser firing, as well as more powerful brain-like computing- and learning capabilities. These design principles are especially suited for creating autonomously learning devices that, like brains, are able to detect and make sense of salient patterns in very high-dimensional multi-modal input streams, and no longer require training with engineered homogeneous data sets in order to become smart.

\paragraph{Acknowledgments}
I would like to thank Yujie Wu and Guozhang Chen for helpful comments.
This research was partially supported by the Human Brain Project (Grant Agreement number 785907) of the European Union, and a grant from Intel. 




\newpage 


\nolinenumbers
\newgeometry{
    a4paper,
    left=3cm,
    right=3cm,
    top=3cm,
    bottom=4cm
}

\begin{center}

~
\\
~
\\

{\LARGE Supplementary Information for the Perspective Article}\\[1em]
{\LARGE "How can neuromorphic hardware attain brain-like functional capabilities?"}\\[1.5em]
{\large Wolfgang Maass}\\
{\large 24th October 2023}
\end{center}

\vspace{2em}

We point here to literature that provides a deeper understanding of the four design principles that are sketched in the article.

\vspace{0.5cm}

\textbf{\em
To Design Principle 1:} \citep{mountcastle1998perceptual} is an important source for information about the structure of the neocortex and cortical microcircuits. It is still very readable and relevant. Further insight into functional specialization of genetically different types of pyramidal cells are provided by \citep{o2021and, musall2023pyramidal}. Experimental evidence for the fact that excitatory neurons (pyramidal cells) are generically under an inhibitory lock, and require disinhibition for firing was provided in \citep{haider2013inhibition}.

Experimental data on gating of synaptic plasticity in CMs are reviewed in \citep{magee2020synaptic, chereau2022circuit}. Data on the role of disinhibition for synaptic plasticity can be found for example in \citep{letzkus2015disinhibition}.
Experimental data on the diversity of dopamine signals were provided by \citep{engelhard2019specialized}. Note that the biologically found gating signals can also be viewed as learning signals for e-prop \citep{bellec2020solution}, and therefore support also some forms of network gradient descent learning that can be implemented in NMHW such as Loihi. 

\vspace{0.5cm}

\textbf{\em To Design Principle 2:} The vision to compute with rank-order coding of single spikes had apparently been first proposed by \citep{thorpe1990spike}. In \citep{maass1994computational} it was rigorously proven that this type of rank-order coding enables spiking neural networks to emulate ANNs with single spikes, rather than firing rates. Apparently the first experimental evidence for the more noise robust biologically found type of rank order coding was provided in \citep{driscoll2017dynamic}. Data from many more brain areas are provided in \citep{koay2022sequential}. The energy efficiency of rank order coding was recently analyzed in \citep{boahen2022dendrocentric}. Properties of CMs that support soft rank order doing and a quantitative measure for rank order coding can be found in \citep{chen2023data}.

\vspace{0.5cm}

\textbf{\em To Design Principle 3:} Functional segregation on the level of single neurons of the neocortex was discussed for example in \citep{houweling2008behavioural, dalgleish2020many}. A quantitative measure for functional segregation was introduced in \citep{chen2023data}, and it was shown that CM-like NN models, but not randomly connected neural networks exhibit segregation and integration of information.

\vspace{0.5cm}

\textbf{\em
To Design Principle 4:} The diversity of synaptic plasticity rules found in the neocortex is reviewed for example in \citep{larsen2015synapse, magee2020synaptic, chereau2022circuit, mcfarlan2023plasticitome}. Specific examples of neural coding properties of neurons that are not likely to emerge through BPTT are discussed for example in \citep{vinje2000sparse, olshausen2004sparse, keller2020feedback, ficsek2023cortico}.

\printbibliography

@article{thorpe1990spike,
  title={Spike arrival times: A highly efficient coding scheme for neural networks},
  author={Thorpe, Simon J},
  journal={Parallel processing in neural systems},
  pages={91--94},
  year={1990},
  publisher={Elsevier}
}

@article{o2023molecularly,
  title={Molecularly targetable cell types in mouse visual cortex have distinguishable prediction error responses},
  author={O’Toole, Sean M and Oyibo, Hassana K and Keller, Georg B},
  journal={Neuron},
  year={2023},
  publisher={Elsevier}
}

@article{rao2022long,
  title={A long short-term memory for AI applications in spike-based neuromorphic hardware},
  author={Rao, Arjun and Plank, Philipp and Wild, Andreas and Maass, Wolfgang},
  journal={Nature Machine Intelligence},
  volume={4},
  number={5},
  pages={467--479},
  year={2022},
  publisher={Nature Publishing Group UK London}
}

@article{tononi1998complexity,
  title={Complexity and coherency: integrating information in the brain},
  author={Tononi, Giulio and Edelman, Gerald M and Sporns, Olaf},
  journal={Trends in cognitive sciences},
  volume={2},
  number={12},
  pages={474--484},
  year={1998},
  publisher={Elsevier}
}

@article{maass1994computational,
  title={On the computational complexity of networks of spiking neurons},
  author={Maass, Wolfgang},
  journal={Advances in neural information processing systems},
  volume={7},
  year={1994}
}

@article{boahen2022dendrocentric,
  title={Dendrocentric learning for synthetic intelligence},
  author={Boahen, Kwabena},
  journal={Nature},
  volume={612},
  number={7938},
  pages={43--50},
  year={2022},
  publisher={Nature Publishing Group UK London}
}

@article{musall2023pyramidal,
  title={Pyramidal cell types drive functionally distinct cortical activity patterns during decision-making},
  author={Musall, Simon and Sun, Xiaonan R and Mohan, Hemanth and An, Xu and Gluf, Steven and Li, Shu-Jing and Drewes, Rhonda and Cravo, Emma and Lenzi, Irene and Yin, Chaoqun and others},
  journal={Nature neuroscience},
  volume={26},
  number={3},
  pages={495--505},
  year={2023},
  publisher={Nature Publishing Group US New York}
}

@article{olshausen2004sparse,
  title={Sparse coding of sensory inputs},
  author={Olshausen, Bruno A and Field, David J},
  journal={Current opinion in neurobiology},
  volume={14},
  number={4},
  pages={481--487},
  year={2004},
  publisher={Elsevier}
}

@article{dalgleish2020many,
  title={How many neurons are sufficient for perception of cortical activity?},
  author={Dalgleish, Henry WP and Russell, Lloyd E and Packer, Adam M and Roth, Arnd and Gauld, Oliver M and Greenstreet, Francesca and Thompson, Emmett J and H{\"a}usser, Michael},
  journal={Elife},
  volume={9},
  pages={e58889},
  year={2020},
  publisher={eLife Sciences Publications, Ltd}
}

@article{o2021and,
  title={Why and how the brain weights contributions from a mixture of experts},
  author={O’Doherty, John P and Lee, Sang Wan and Tadayonnejad, Reza and Cockburn, Jeff and Iigaya, Kyo and Charpentier, Caroline J},
  journal={Neuroscience \& Biobehavioral Reviews},
  volume={123},
  pages={14--23},
  year={2021},
  publisher={Elsevier}
}

@article{letzkus2015disinhibition,
  title={Disinhibition, a circuit mechanism for associative learning and memory},
  author={Letzkus, Johannes J and Wolff, Steffen BE and L{\"u}thi, Andreas},
  journal={Neuron},
  volume={88},
  number={2},
  pages={264--276},
  year={2015},
  publisher={Elsevier}
}

@article{zador2019critique,
  title={A critique of pure learning and what artificial neural networks can learn from animal brains},
  author={Zador, Anthony M},
  journal={Nature communications},
  volume={10},
  number={1},
  pages={3770},
  year={2019},
  publisher={Nature Publishing Group UK London}
}

@article{stockl2021structure,
  title={Structure induces computational function in networks with diverse types of spiking neurons},
  author={St{\"o}ckl, Christoph and Lang, Dominik and Maass, Wolfgang},
  journal={bioRxiv},
  pages={2021--05},
  year={2021},
  publisher={Cold Spring Harbor Laboratory}
}

@article{chen2022data,
  title={A data-based large-scale model for primary visual cortex enables brain-like robust and versatile visual processing},
  author={Chen, Guozhang and Scherr, Franz and Maass, Wolfgang},
  journal={Science Advances},
  volume={8},
  number={44},
  pages={eabq7592},
  year={2022},
  publisher={American Association for the Advancement of Science}
}

@article{engelhard2019specialized,
  title={Specialized coding of sensory, motor and cognitive variables in VTA dopamine neurons},
  author={Engelhard, Ben and Finkelstein, Joel and Cox, Julia and Fleming, Weston and Jang, Hee Jae and Ornelas, Sharon and Koay, Sue Ann and Thiberge, Stephan Y and Daw, Nathaniel D and Tank, David W and others},
  journal={Nature},
  volume={570},
  number={7762},
  pages={509--513},
  year={2019},
  publisher={Nature Publishing Group UK London}
}

@article{billeh2020systematic,
  title={Systematic integration of structural and functional data into multi-scale models of mouse primary visual cortex},
  author={Billeh, Yazan N and Cai, Binghuang and Gratiy, Sergey L and Dai, Kael and Iyer, Ramakrishnan and Gouwens, Nathan W and Abbasi-Asl, Reza and Jia, Xiaoxuan and Siegle, Joshua H and Olsen, Shawn R and others},
  journal={Neuron},
  year={2020},
  publisher={Elsevier}
}

@article{harris_sheperd,
author = {Harris, Kenneth and Shepherd, Gordon},
year = {2015},
month = {02},
pages = {170-181},
title = {The neocortical circuit: Themes and variations},
volume = {18},
journal = {Nature neuroscience},
doi = {10.1038/nn.3917}
}

@article{bellec2020solution,
  title={A solution to the learning dilemma for recurrent networks of spiking neurons},
  author={Bellec, Guillaume and Scherr, Franz and Subramoney, Anand and Hajek, Elias and Salaj, Darjan and Legenstein, Robert and Maass, Wolfgang},
  journal={Nature communications},
  volume={11},
  number={1},
  pages={3625},
  year={2020},
  publisher={Nature Publishing Group UK London}
}

@book{mountcastle1998perceptual,
  title={Perceptual neuroscience: The cerebral cortex},
  author={Mountcastle, Vernon B},
  year={1998},
  publisher={Harvard University Press}
}

@article{koay2022sequential,
  title={Sequential and efficient neural-population coding of complex task information},
  author={Koay, Sue Ann and Charles, Adam S and Thiberge, Stephan Y and Brody, Carlos D and Tank, David W},
  journal={Neuron},
  volume={110},
  number={2},
  pages={328--349},
  year={2022},
  publisher={Elsevier}
}

@article{houweling2008behavioural,
  title={Behavioural report of single neuron stimulation in somatosensory cortex},
  author={Houweling, Arthur R and Brecht, Michael},
  journal={Nature},
  volume={451},
  number={7174},
  pages={65--68},
  year={2008},
  publisher={Nature Publishing Group}
}

@article{driscoll2017dynamic,
  title={Dynamic reorganization of neuronal activity patterns in parietal cortex},
  author={Driscoll, Laura N and Pettit, Noah L and Minderer, Matthias and Chettih, Selmaan N and Harvey, Christopher D},
  journal={Cell},
  volume={170},
  number={5},
  pages={986--999},
  year={2017},
  publisher={Elsevier}
}

@article{chen2023data,
  title={Data-based large-scale models provide a window into the organization of cortical computations},
  author={Chen, Guozhang and Scherr, Franz and Maass, Wolfgang},
  journal={bioRxiv},
  pages={2023--04},
  year={2023},
  publisher={Cold Spring Harbor Laboratory}
}

@article{keller2020feedback,
  title={Feedback generates a second receptive field in neurons of the visual cortex},
  author={Keller, Andreas J and Roth, Morgane M and Scanziani, Massimo},
  journal={Nature},
  volume={582},
  number={7813},
  pages={545--549},
  year={2020},
  publisher={Nature Publishing Group UK London}
}

@article{vinje2000sparse,
  title={Sparse coding and decorrelation in primary visual cortex during natural vision},
  author={Vinje, William E and Gallant, Jack L},
  journal={Science},
  volume={287},
  number={5456},
  pages={1273--1276},
  year={2000},
  publisher={American Association for the Advancement of Science}
}

@article{douglas2004neuronal,
  title={Neuronal circuits of the neocortex},
  author={Douglas, Rodney J and Martin, Kevan AC},
  journal={Annu. Rev. Neurosci.},
  volume={27},
  pages={419--451},
  year={2004},
  publisher={Annual Reviews}
}

@article{mcfarlan2023plasticitome,
  title={The plasticitome of cortical interneurons},
  author={McFarlan, Amanda R and Chou, Christina YC and Watanabe, Airi and Cherepacha, Nicole and Haddad, Maria and Owens, Hannah and Sj{\"o}str{\"o}m, P Jesper},
  journal={Nature Reviews Neuroscience},
  volume={24},
  number={2},
  pages={80--97},
  year={2023},
  publisher={Nature Publishing Group UK London}
}

@article{larsen2015synapse,
  title={Synapse-type-specific plasticity in local circuits},
  author={Larsen, Rylan S and Sj{\"o}str{\"o}m, P Jesper},
  journal={Current opinion in neurobiology},
  volume={35},
  pages={127--135},
  year={2015},
  publisher={Elsevier}
}

@article{haider2013inhibition,
  title={Inhibition dominates sensory responses in the awake cortex},
  author={Haider, Bilal and H{\"a}usser, Michael and Carandini, Matteo},
  journal={Nature},
  volume={493},
  number={7430},
  pages={97--100},
  year={2013},
  publisher={Nature Publishing Group UK London}
}

@article{magee2020synaptic,
  title={Synaptic plasticity forms and functions},
  author={Magee, Jeffrey C and Grienberger, Christine},
  journal={Annual review of neuroscience},
  volume={43},
  pages={95--117},
  year={2020},
  publisher={Annual Reviews}
}

@inproceedings{chereau2022circuit,
  title={Circuit mechanisms for cortical plasticity and learning},
  author={Ch{\'e}reau, Ronan and Williams, Leena E and Bawa, Tanika and Holtmaat, Anthony},
  booktitle={Seminars in cell \& developmental biology},
  volume={125},
  pages={68--75},
  year={2022},
  organization={Elsevier}
}

@article{ficsek2023cortico,
  title={Cortico-cortical feedback engages active dendrites in visual cortex},
  author={Fi{\c{s}}ek, Mehmet and Herrmann, Dustin and Egea-Weiss, Alexander and Cloves, Matilda and Bauer, Lisa and Lee, Tai-Ying and Russell, Lloyd E and H{\"a}usser, Michael},
  journal={Nature},
  pages={1--8},
  year={2023},
  publisher={Nature Publishing Group UK London}
}

@article{campagnola2022local,
  title={Local connectivity and synaptic dynamics in mouse and human neocortex},
  author={Campagnola, Luke and Seeman, Stephanie C and Chartrand, Thomas and Kim, Lisa and Hoggarth, Alex and Gamlin, Clare and Ito, Shinya and Trinh, Jessica and Davoudian, Pasha and Radaelli, Cristina and others},
  journal={Science},
  volume={375},
  number={6585},
  pages={eabj5861},
  year={2022},
  publisher={American Association for the Advancement of Science}
}

\end{document}